# SEvoBench : A C++ Framework For Evolutionary Single-Objective Optimization Benchmarking

Yongkang Yang
School of Science
University of Science And Technology Liaoning
Anshan, China
yangyongkang@ustl.edu.cn

Jian Zhao
School of Science
University of Science And Technology Liaoning
Anshan, China
zhao@ustl.edu.cn

Tengfei Yang
School of Science
University of Science And Technology Liaoning
Anshan, China
232070100042@stu.ustl.edu.cn

## ABSTRACT

We present SEvoBench, a modern C++ framework for evolutionary computation (EC), specifically designed to systematically benchmark evolutionary single-objective optimization algorithms. The framework features modular implementations of Particle Swarm Optimization (PSO) and Differential Evolution (DE) algorithms, organized around three core components: (1) algorithm construction with reusable modules, (2) efficient benchmark problem suites, and (3) parallel experimental analysis. Experimental evaluations demonstrate the framework's superior performance in benchmark testing and algorithm comparison. Case studies further validate its capabilities in algorithm hybridization and parameter analysis. Compared to existing frameworks, SEvoBench demonstrates three key advantages: (i) highly efficient and reusable modular implementations of PSO and DE algorithms, (ii) accelerated benchmarking through parallel execution, and (iii) enhanced computational efficiency via SIMD (Single Instruction Multiple Data) vectorization for large-scale problems.

• **Computing methodologies** → Continuous space search.

## 1 Introduction

Single-objective optimization algorithms form the foundation for studying more complex variants, including multi-objective and constrained optimization methods. Evolutionary algorithms (EAs) are widely used to solve single-objective continuous black-box optimization problems (SOPs), especially when the problems lack explicit mathematical expressions, or are non-differentiable, multimodal, noisy, etc.

A robust benchmarking framework for EC enhances the reproducibility of experiments and facilitates the improvement and dissemination of algorithms. Currently, there are several leading benchmarking frameworks in this field, focusing on two main aspects of design. One aspect focuses on designing benchmark suites and evaluating algorithm performance. For example, Hansen et al. designed the COCO [1] framework for comparing continuous black-box optimization algorithms. For SOPs, COCO includes the BBOB benchmark suite [20], which consists of 24 complex optimization problems. COCO evaluates runtime by simulating the Conceptual Restart Algorithm through running different problem instances for each problem in the suite. COCO primarily uses empirical cumulative distribution function (ECDF) curves for performance evaluation. Another well-known benchmarking framework is IOHexperimenter [2] which provides the BBOB suite and offers user-friendly interfaces for algorithms and loggers to facilitate testing. The data collected from experiments can be evaluated using ECDF curves.

The other aspect is providing frameworks for EAs. One aspect focuses on the modular implementation of EAs. This means that an instance of EA is constructed by the combination of multiple modules. Paradiseo [3] introduced the concept of modular design for algorithms at an early stage and provided the EO module, which integrates modular implementations of genetic algorithm and PSO algorithm [4]. In recent years, as EC has advanced, researchers have identified key modules through empirical studies for designing specific EAs. For the DE algorithm [5], *Modular DE*[1] [6] provides the necessary modules to be considered when implementing the DE. Similarly, for the PSO, PSO-X [7] defines the modules to be considered in the design of PSO. At the software level, the pso-de-framework [2] [8,9] offers a well-structured modular implementation for both DE and PSO. In recent years, other frameworks have focused on specific areas of EAs. For example, the PyPop7 [10] framework focuses on large-scale SOPs testing and integrates many EAs; PlatEMO [11] collects many multi-objective algorithms, while MEALPY [12] gathers a wide range of metaphor-based algorithms. EvoX [13] and evosax [23] use GPU technology to enhance the efficiency of EAs.

---
[1] https://github.com/Dvermetten/ModDE

Inspired by state-of-the-art frameworks in EC [2,3,6,7,8,9], we propose SEvoBench [3], an open-source C++20 framework for single-objective optimization benchmarking. This framework introduces novel designs in the following aspects. At the algorithmic level, it integrates the DE module and PSO module. The DE module, inspired by *Modular DE*, incorporates a broader set of modular components compared to pso-de-framework while also benefiting from better design patterns. This results in improved scalability and reusability compared to the software implementation of *Modular DE*. The PSO module shares similar modular components with pso-de-framework and Paradiso but achieves higher computational efficiency.

Regarding the benchmark suite, the legacy C++ implementation [4] of the CEC suite adopts a C-style approach, which violates Modern C++ principles through design flaws including: (1) global variable pollution, (2) manual memory management, (3) thread-unsafety, and (4) limited reusability/extensibility. IOHexperimenter employs the CRTP (Curiously Recurring Template Pattern) to implement CEC2022, enhancing reusability and extensibility. SEvoBench further advances this by redesigning CEC2014 and CEC2017, introducing several new features. Specifically, each problem is designed to generate multiple random instances, like the approach used in BBOB suite.

At the experiment level, IOHexperimenter's ability to provide diverse metrics for tracking algorithm performance stems from its extensible logger component. SEvoBench also offers an extensible logger component, supporting the tracking of performance metrics that meet CEC competition requirements, while enabling parallelization of the testing process. Notably, the C++ kernel version of IOHexperimenter does not support parallelism.

In terms of high-performance computing, for large-scale SOPs, SIMD technology is introduced to enhance computational efficiency.

The rest of this paper is organized as follows. Section 2 introduces the architecture and modules of the framework. Section 3 presents novel techniques introduced in the framework through experimental studies. Section 4 presents two case studies of the framework's application. Section 5 concludes our paper and discusses future work.

## 2 THE SEVOBENCH FRAMEWORK

SEvoBench is implemented in C++20 standard, which makes extensive use of template programming techniques. The framework focuses on basic benchmarking of SOPs. This section will briefly describe the architecture of the framework and then go into more detail about the modules.

### 2.1 Architecture

The architecture consists of the three core modules: the Algorithm module, the Problem module, and the Experiment module. The Algorithm module includes built-in algorithms and supports user-defined algorithms. The Problem module contains built-in suites and supports user-defined suites. The Experiment module evaluates EAs against benchmark suites.

### 2.2 Algorithm Module

The Algorithm Module is an important part of the framework. Several EAs are pre-implemented in the framework and user-defined algorithms can also be added to the framework.

1) User's Algorithm: The SEvoBench framework requires user-defined EAs to be implemented as callable objects that: (1) accept a fitness evaluation function as their primary parameter, and (2) fully encapsulate the evolutionary optimization logic. These callable objects must conform to the interface specification of the *evo_bench* function (in the Experiment module), ensuring proper integration with the benchmarking infrastructure while maintaining algorithmic flexibility. As in IOHexperimenter, the passed algorithm class can be considered a function object.

2) Built-in Algorithm: In the modular implementation of EAs, at the software level, three conditions should be met: (1) inclusion of essential modules, (2) ease of extension and replacement of individual modules, with the ability to extend or replace modules without affecting the stability of the algorithm framework, and (3) the algorithm framework should possess good reusability, allowing users to apply different termination conditions to implement restart without the need to copy/paste the algorithm code into another function. SEvoBench provides DE and PSO modules, with Figures 1 and 2 illustrating the UML design diagrams of the DE and PSO modules, respectively. We have employed the Strategy pattern, a design pattern also used by Paradiseo and pso-de-framework. This design pattern is well-suited for DE and PSO, where the overall algorithm framework remains stable, and the modules of the algorithm are subject to change. The use of the Strategy pattern effectively satisfies condition (2). In the DE module, the composition of the modules not only includes all the modules required by *Modular DE*, but also introduces a base class for external archiving [19], virtual functions encapsulate the archiving behavior, which is absent in *Modular DE*. Additionally, the Python implementation of *Modular DE* does not use the Strategy pattern; instead, it uses 'if-else' statements for extending and replacing modules. In the pso-de-framework, the external archiving mechanism and population reduction mechanism have not been considered. For simplified construction, the Builder pattern is introduced. Within the UML model, the generic type *T* is constrained to floating-point types. In SEvoBench, this is enforced via the *std::floating_point* concept (C++20), restricting *T* to *float*, *double,* or *long double* at compile time. In Listing 1, the code demonstrates the construction of modular PSO and DE using the Builder pattern. The code exemplifies the modular construction of two distinct EAs using SEvoBench's Builder pattern: (1) a PSO variant employing spherical velocity updates [24] with ring (*lbest*) topology, and (2) a canonical Success-History based Adaptive DE [14] (SHADE) algorithm without external archive. It should be noted that while the framework employs *unique_ptr* for automatic memory management of algorithmic modules during construction, users retain the ability to access and manipulate these modules through raw pointers obtained via member functions of the instantiated algorithm class. Beyond the DE and PSO modules, SEvoBench provides native implementations of additional EAs[5].

---

[2] https://github.com/rickboks/pso-de-framework
[3] https://github.com/yangyongkang2000/SEvoBench
[4] https://github.com/yangyongkang2000/SEvoBench/blob/main/tests/cec22_test_func.cpp

```cpp
auto topo=make_unique<lbest_topology<float>>();
auto upda=make_unique<spherical_update<float>>();
auto pso=pso_algorithm_builder().
        topology(move(topo)).
        update(move(upda)).
        build();
auto p=make_unique<shade_parameter<float>>();
auto m=make_unique<ttpb1_mutation<float>>();
auto c=make_unique<binomial_crossover<float>>();
auto h=make_unique<midpoint_target_repair<float>>();
auto s=make_unique<de_population<float>>();
auto shade=de_algorithm_builder().
        mutation(move(m)).
        parameter(move(p)).
        population_strategy(move(s)).
        crossover(move(c)).
        constraint_handler(move(h)).
        build();
```

**Listing 1: Builder pattern implementation for modular PSO and DE algorithm**

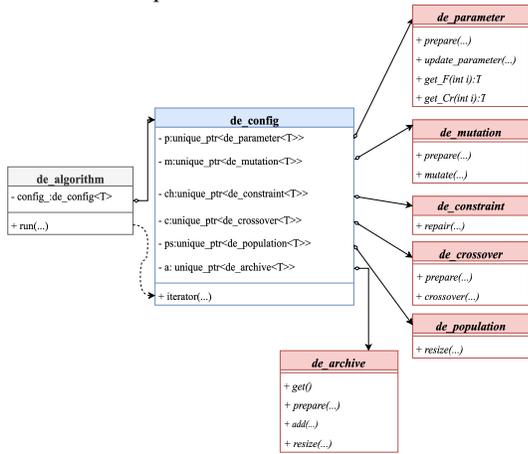

**Figure 1: UML diagram of DE module**

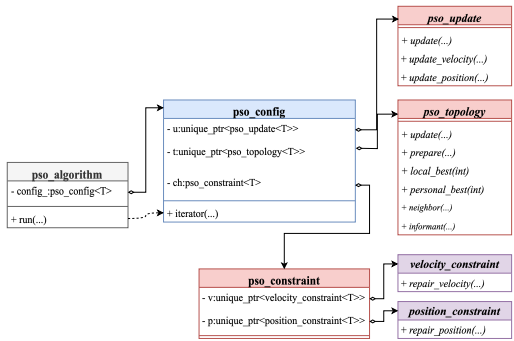

**Figure 2: UML diagram of PSO module**

---

[5] https://github.com/yangyongkang2000/SEvoBench/tree/main/include/SEvoBench/algorithm/other_algorithm

### 2.3 Problem Module

The Problem module is an essential part of the framework, which provides the problems required for algorithm benchmarking. Like the Algorithm module, The Problem module has a built-in set of benchmark suites for SOPs and supports user-defined problems.

We demonstrate the design patterns and techniques used in the problem modules by implementing the CEC2017 benchmark suite. Figure 3 shows the UML diagram of the class *cec2017*. The CEC2017 implementation employs the CRTP pattern to reuse data and member functions from the class *cec_common*. This approach is similarly adopted in IOHexperimenter's CEC2022 suite. On the software level, each problem class is an independent type of object, and a benchmark suite serves as a container that stores multiple problems. To facilitate the storage and invocation of multiple problem types, we apply type erasure (e.g., via virtual functions) to unify their interfaces. SEvoBench implements type erasure through virtual functions, while IOHexperimenter uses a similar technique. The construction of the benchmark suite involves the use of the Singleton pattern combined with the Factory pattern in IOHexperimenter, while SEvoBench employs template metaprogramming techniques combined with the Builder pattern. Listing 2 compares the benchmark suite construction mechanisms in SEvoBench and IOHexperimenter.

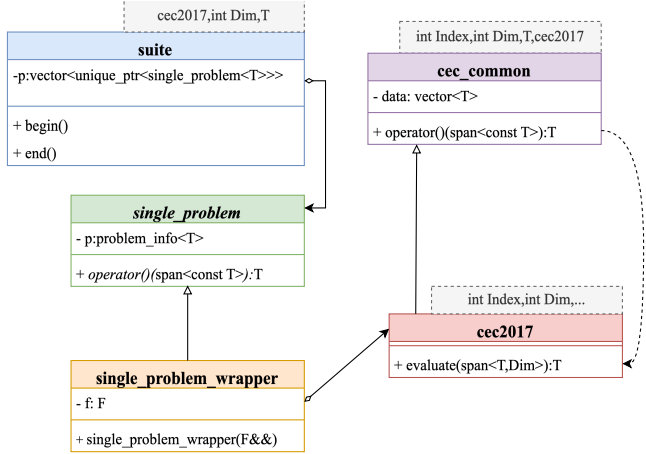

Figure 3: UML diagram of *cec2017*

## 2.4 Experiment Module

When the framework includes a series of algorithms and benchmark suites, we provide the Experiment module to evaluate the performance of the algorithms on the benchmark suites. The Experiment module provides a function interface for benchmarking different algorithms on different benchmark suites. In Listing 3, the *evo_bench* function connects the Algorithm module and the Problem module. The *evo_bench* function accepts four parameters. The first is the function object representing the EA, the second is the benchmark suite object, the third parameter *obs* is a universal reference (*auto&&*) that accepts any polymorphic subclass of *suite_observer*, which is used to track the performance of the optimization process, and the fourth is the number of independent runs. The class *suite_observer*, like the *logger* in IOHexperimenter, uses the Observer pattern, where it decides whether to record relevant data each time a problem's fitness is evaluated. However, in IOHexperimenter, the problem and logger are tightly coupled, while in SEvoBench, the *evo_bench* function internally constructs a *suite_problem* wrapper (as shown in Figure 4) that combines a polymorphic reference (*suite_observer &*) to the observer with a polymorphic pointer (*single_problem**) to the problem instance, this composition is intentionally encapsulated within the function's implementation.

In SEvoBench, we define the class *best_so_far_record* as a subclass of *suite_observer* to record the best solution so far, as required by the CEC competition. Moreover, compared to the container used in IOHexperimenter's class *Store* for recording metrics, which does not consider thread safety, the *best_so_far_record* class features a storage container that is pre-allocated during the construction phase, ensuring thread safety. Notably, the template parameter *parallel* controls whether the test process is parallel or sequential, when *parallel* is true, the test tasks are distributed among the threads in the thread pool, and each thread runs its tasks asynchronously, achieving parallelism.

```
// IOHexperimenter
SuiteRegistry<RealSingleObjective>::instance()
.create("BBOB",problems,instances, dimensions);
// SEvoBench
suite_builder<cec2017>()
    .dim<Dim>()
    .problem_index(problems)
    .instance_count(1)
    .build();
```

Listing 2: Benchmark suite construction in SEvoBench and IOHexperimenter

```
template <bool parallel = true>
void evo_bench(auto &&alg, auto &&su, auto
&&obs,int independent_runs);
```

Listing 3: Declaration of the *evo_bench* function

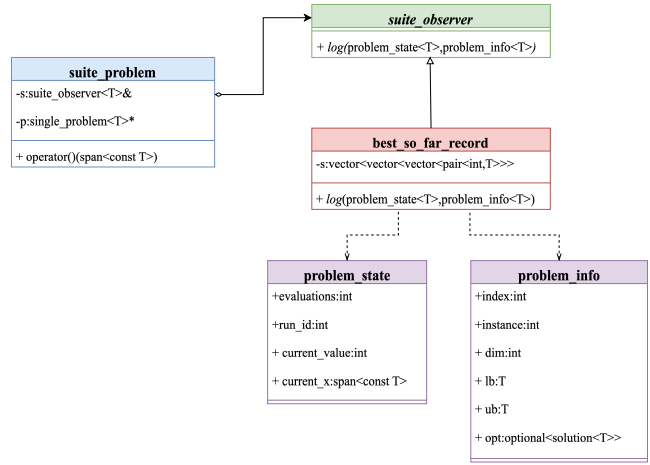

Figure 4: UML class diagram of SEvoBench's core monitoring components

## 3 EXPERIMENTAL STUDY

In this section, three experiments are conducted to verify: (1) the efficient EA implementation in the SEvoBench framework, (2) the reduction of testing time through parallel computing, and (3) the contribution of the introduced acceleration techniques to large-scale benchmarking. Unless otherwise specified, all experiments in this section were performed on a Mac OS 13.7.6 system with an Intel Core i5-7360U processor and 8GB memory, using Python 3.9.6 and compiled with Clang version 21.0.0. The implementation code for these experiments is publicly accessible in the project's GitHub repository[6].

### 3.1 Runtime efficiency comparison of EAs across frameworks

In this section, we conduct a comparative analysis of EA implementation efficiency across three EC frameworks: Paradiseo v3.0.0 (C++), PyPop7 v0.0.81 (Python), and EvoX v0.9.0 (Python). For PSO algorithms, uniform parameter configurations were established using the PSO module (SEvoBench) and EO module (Paradiseo), with experimental parameters set to a population size of 30, maximum iteration limit of 1,000, and 10 independent trials to mitigate stochastic variance. Computational performance was evaluated using the sphere function benchmark, with mean execution time aggregated across all runs.

To extend the comparison to DE paradigms, we implemented SHADE via the DE module in SEvoBench, benchmarking it against native SHADE implementations in PyPop7 and EvoX. This experimental setup adopted a population size of 100 individuals subject to a computational budget constraint of 30,000 function evaluations, with 10 independent trials executed to ensure statistical reliability. Framework-level efficiency metrics were calculated from mean execution times, using the 30-dimensional Rosenbrock function as the benchmark problem.

Figure 5 illustrates the computational efficiency comparison between SEvoBench and three EC frameworks using bar charts. The comparative analysis with Paradiseo, demonstrates significant performance advantages of SEvoBench. The benchmark results reveal: (i) SEvoBench (v1) utilizing the framework's default random number generator achieves a 4× speedup over Paradiseo. (ii) SEvoBench (v2) employing the same random number generator as Paradiseo maintains a 70% performance advantage. Our experimental results demonstrate that SEvoBench exhibits approximately two orders of magnitude higher computational efficiency than both EvoX and PyPop7 frameworks. This performance disparity primarily stems from fundamental architectural differences: SEvoBench's C++ implementation benefits from native code compilation and explicit SIMD optimization, while Python's interpreted execution model inherently limits PyPop7's performance. The EvoX framework presents a more complex performance profile, as documented in prior work [13]. While EvoX achieves three orders of magnitude acceleration when leveraging NVIDIA GPUs through JAX's just-in-time compilation, our controlled experiments reveal this advantage diminishes significantly on CPU-only systems. Specifically, when JAX defaults to its CPU backend, EvoX's performance degrades to levels below PyPop7's NumPy-based implementation, particularly on processors with integrated graphics.

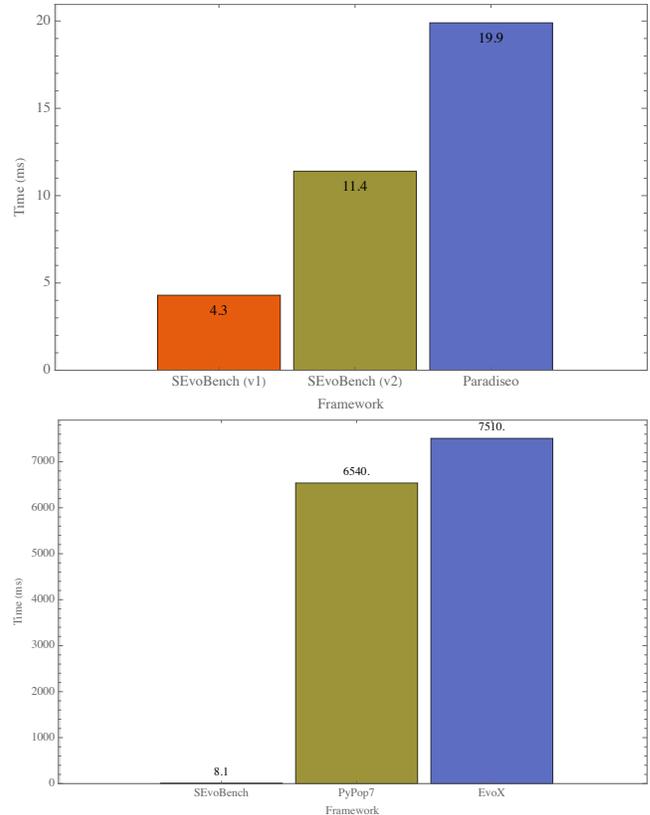

**Figure 5: Comparison of runtime performance across frameworks**

### 3.2 Computational efficiency comparison with IOHexperimenter

This section presents a comparative runtime analysis between SEvoBench and the C++-based IOHexperimenter (v0.3.18) framework for EA testing. We employ DE with a population size of 100 and a maximum evaluation budget of 200,000 as the benchmark algorithm. To ensure statistical significance, each test problem from the CEC2022 benchmark suite (20-dimensional) is executed 30 times independently. Crucially, identical DE algorithm code generated by DeepSeek-V3 [15] is utilized across both frameworks to enable fair comparison. The LLM-generated code was carefully verified and refined through expert review and empirical testing to ensure correctness before being adopted for benchmarking.

During the testing process, the best-so-far fitness value is recorded every 200 fitness evaluations. In SEvoBench, the statistical results are stored using the class *best_so_far_recorder* within the Experiment module. Similarly, in IOHexperimenter, the results are logged using the class *Store* from the logger module.

The bar chart in Figure 6 reveals that SEvoBench attains a threefold improvement (over 3× higher) in computational efficiency compared to IOHexperimenter for DE execution. Through detailed hotspot analysis using performance profiling tools, we attribute this performance gap to SEvoBench's architectural optimizations in both memory management and parallel execution support.

---

[6] https://github.com/yangyongkang2000/SEvoBench/tree/main/example/benchmark

(1) Memory management optimization: SEvoBench predefines the dimensionality of the test suite as a template parameter, enabling transformed solution vectors to be stored directly in stack memory during fitness evaluations. This design eliminates runtime memory allocation overhead. In contrast, IOHexperimenter allocates and deallocates heap memory for transformed solution vectors during each fitness function call, incurring significant latency under high-frequency invocations (e.g., 200,000 evaluations).

(2) Parallel execution: SEvoBench's class *best_so_far_record* implements an asynchronous storage mechanism for statistical metrics, decoupling data logging from computation threads. Conversely, IOHexperimenter's class *Store* employs a nested hash table structure with synchronous operations, which fundamentally prevents parallel execution as its core design lacks thread-safety mechanisms. Unlike SEvoBench's native parallel support, IOHexperimenter's C++ kernel provides no parallelism capabilities, forcing all evaluations to proceed sequentially.

Furthermore, SEvoBench provides native parallel execution support that fully utilizes modern multi-core architectures. When enabled, this yields an additional 2.5× speedup (measured on a 2-core/4-thread processor), whereas IOHexperimenter's single-threaded architecture and non-asynchronous data structures fundamentally prohibit parallel execution.

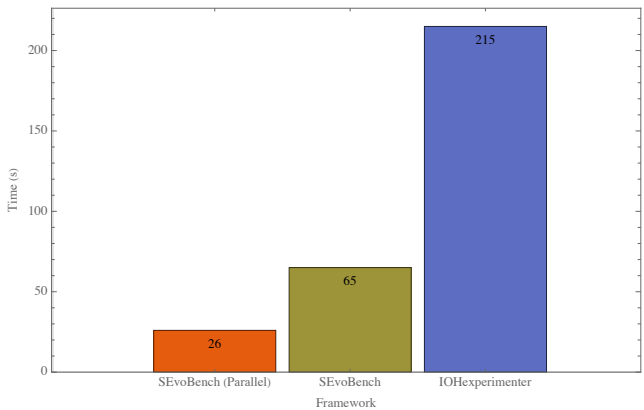

**Figure 6. Computational time comparison of SEvoBench and IOHexperimenter**

### 3.3 Acceleration technology

For large-scale SOPs, the high-dimensional nature leads to high computational complexity and low efficiency. Therefore, SEvoBench introduces SIMD instructions to enhance performance. When using the SIMD instruction set, the CPU utilizes vector registers to access and process multiple data elements simultaneously, enabling vectorized computations. We utilize the open-source SIMD library, *vectorclass*[7], which includes a wide range of SIMD mathematical functions, fully leveraging the performance advantages of modern processors.

In this study, we integrate SIMD vectorization into the large-scale SOPs benchmark suite CEC2010. All subsequent experiments were performed on a dedicated server running CentOS Linux 7.9.2009 (Core) with GNU GCC 10.5.0, featuring dual Intel® Xeon® Platinum 8358 processors (2.60GHz) and 342GB of available system memory. Our experimental design comprises two distinct configurations.

Group 1: A specialized benchmark subset is constructed by selecting CEC2010 problems requiring intensive trigonometric/exponential function computations (mathematical operations benefiting significantly from SIMD acceleration).

Group 2: Comprehensive benchmarking is conducted using the complete CEC2010 test suite. We compare wall time between an explicit SIMD-optimized implementation using *vectorclass* and a compiler-auto-vectorized implementation with *-O3 -march=native* flags.

The Competitive Swarm Optimizer (CSO) [16] is employed with a population size of 300 and a maximum fitness evaluation budget of 1,000,000. Each 1000-dimensional problem instance is executed 5 times under single-instance configuration. Runtime performance is statistically compared across computational paradigms. The experiments employ single-precision floating-point arithmetic (float type) because vector registers accommodate different numbers of elements depending on precision. This precision-dependent capacity necessitates implementing the floating-point type as a template parameter.

The figure 7 illustrates the time required for different computational paradigms, denoted by a two-letter code "XX". The first letter indicates whether parallelization is enabled ("P" for parallel, "N" for serial), and the second letter indicates whether SIMD optimization is applied ("S" for SIMD-enabled, "N" for SIMD-disabled). Notably, the SIMD optimization in this context refers to the use of the *vectorclass* library to accelerate specific mathematical functions, such as trigonometric and exponential functions. However, for simpler problems like the sphere function, modern compilers (e.g., GCC, Clang) automatically apply SIMD optimizations, reducing the need for explicit library usage.

From the chart, it is evident that the "PS" paradigm (parallel with SIMD) achieves more than a 4× speedup compared to "PN" (parallel without SIMD). Similarly, the "NS" paradigm (serial with SIMD) demonstrates over a 4× speedup compared to "NN" (serial without SIMD). These results highlight the significant performance improvements enabled by SIMD acceleration in fitness function evaluations, substantially reducing the time required to complete a full algorithmic experiment.

Figure 8 presents the experimental results for Group 2, where the performance improvement of the "PS" computational paradigm over "PN" is approximately 79% when evaluated on the entire CEC2010 benchmark test suite. This marginal gain is attributed to the selective application of the *vectorclass* library, which accelerates only those problems involving computationally intensive mathematical functions, such as trigonometric and exponential operations. For the remaining problems, the compiler's inherent SIMD optimization capabilities suffice to achieve near-optimal performance. However, in scenarios like the CEC2013 large-scale benchmark suite, where nearly all solution vector transformations involve trigonometric and exponential function computations [21], the utilization of SIMD libraries yields significant performance improvements. This highlights the critical role of SIMD acceleration in large-scale SOPs, particularly those dominated by complex mathematical operations. The results underscore the importance of tailored SIMD implementations in EC frameworks to maximize computational efficiency.

---
[7] https://github.com/vectorclass/version2

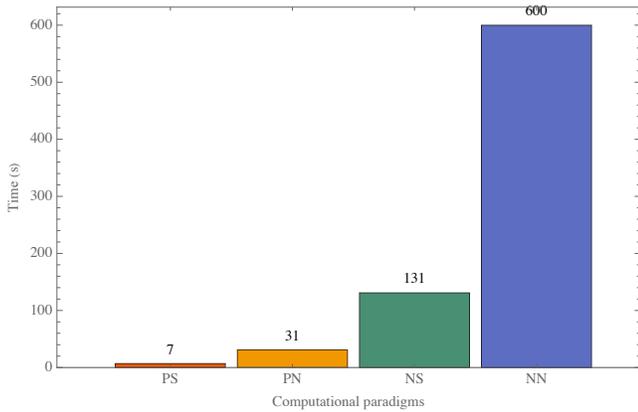

**Figure 7. Comparison of computational time under different paradigms in Group 1**

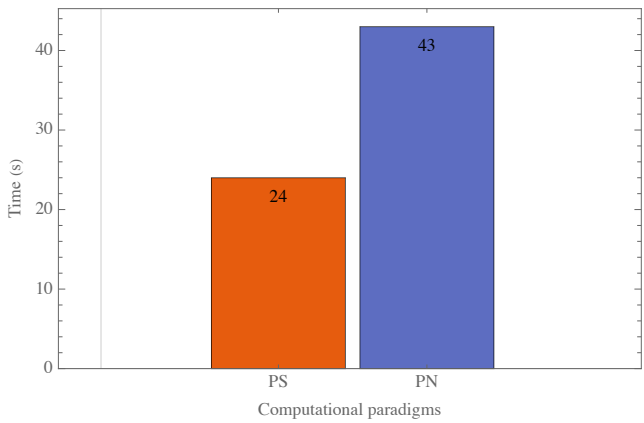

**Figure 8. Comparison of computational time under different paradigms in Group 2**

## 4  CASE STUDY

This section demonstrates the code reusability and extensibility of the SEvoBench framework through two concrete case studies.

The first case study involves replicating the PSODE algorithm from the pso-de-framework. For demonstration purposes, we implemented PSODE with the pairwise/3 selection operator [9]. Listing 4 illustrates the PSODE replication using SEvoBench, where the construction processes of both PSO and SHADE components have been previously shown in Listing 1. This implementation showcases SEvoBench's capability to seamlessly integrate different evolutionary paradigms while maintaining clean architectural separation. Our modular design also enables straightforward implementation of advanced variants like L-SHADE-Restart [8] [22], demonstrating flexible support for both custom stopping criteria and state-triggered restart mechanisms while maintaining full compatibility with base algorithm components.

The second case study extends the functionality of the class *suite_observer* to track the dynamically changing parameters $\mu_F$ and $\mu_{CR}$ in JADE [17]. With assistance from DeepSeek-V3, we implemented this extension by designing a new class structure inspired by SEvoBench's class *best_so_far_record* implementation. The developed module logs these adaptation parameters throughout optimization and exports structured data for analysis, with full implementation available in our repository[9]. Based on the exported data, Figure 9 presents the evolutionary trajectories of the adaptation parameters $\mu_F$ and $\mu_{CR}$ throughout the optimization process, displaying their mean values at each iteration.

## 5  CONCLUSION AND FUTURE WORK

This paper presented SEvoBench, a modern C++20 framework for evolutionary single-objective optimization benchmarking that advances the state-of-the-art through three key innovations. First, our modular design employing Strategy and Builder patterns enables flexible algorithm composition while maintaining performance, demonstrated by 4× speedups over Paradiseo in PSO execution. Second, the redesigned benchmarking infrastructure resolves critical limitations of legacy implementations through thread-safe memory management and parallel evaluation capabilities, achieving 3× faster execution than IOHexperimenter, even in single-threaded scenarios. Third, SIMD vectorization via the *vectorclass* library accelerates large-scale SOPs by 4× speedup on computationally intensive functions. These architectural advantages position SEvoBench as both a practical tool for experimental research and a foundation for future methodological development.

While SEvoBench demonstrates significant advantages in computational efficiency and parallelization, we acknowledge the pioneering contributions of existing frameworks and identify key areas for improvement through cross-framework synergy. First, we recognize that IOHexperimenter remains the gold standard for comprehensive performance analysis—its logger module supports richer metric tracking, COCO-compatible data export, and seamless integration with IOHanalyzer [18] for visualization. Future versions of SEvoBench will adopt these benchmarking best practices while maintaining our architectural advantages in parallel execution. Second, while SEvoBench's DE module successfully implements the core components proposed in *Modular DE*, our current PSO module has not yet fully incorporated all modular design elements suggested in PSO-X. Completing this implementation will strengthen the framework's theoretical foundation and practical utility.

---

[8] https://github.com/yangyongkang2000/SEvoBench/tree/main/example/restart_lshade
[9] https://github.com/yangyongkang2000/SEvoBench/tree/main/example/parameter_observer

```cpp
template <std::floating_point T>
void hybrid_pso_de(auto &&pso_alg, auto &&de_alg,
auto &&pop, auto &&f, T lb, T ub, auto &&alg) {
    using namespace sevobench;
    using namespace sevobench::pso_module;
    auto ps = pop.pop_size();
    auto dim = pop.dim();
    population<T> pop1(ps,dim);
    population<T> pop2(ps,dim);
    population<T> pop3(ps,dim);
    pso_velocity<T> vec(ps,std::vector<T>(dim));
    for (auto &_ : pop)
        _.evaluate(f);
    alg.add_fes(pop.pop_size());
    pso_alg.topology()->prepare(pop);
    do {
        pop1 = pop;
        pop3 = pop;
        pso_alg.iterator(pop1,vec,f,lb,ub,alg);
        de_alg.iterator(pop3,pop2,f,lb,ub,alg);
        for (int i = 0; i < ps; i++)
            for (int j = 0; j < dim; j++)
                vec[i][j] = pop3[i][j] - pop[i][j];
        for (int i = 0; i < ps; i++)
            std::swap(pop[i],
                      pop1[i].fitness() <
            pop3[i].fitness() ? pop1[i] : pop3[i]);
    } while (alg.current_fes() < alg.max_fes());
}

using namespace sevobench;
constexpr int dim = 30;
constexpr int pop_size = 100;
constexpr int max_fes = 1000 * dim;
constexpr float lb = float(-100);
constexpr float ub = float(100);

evolutionary_algorithm alg(max_fes, pop_size, dim);
population<float> pop(pop_size, dim, lb, ub);

hybrid_pso_de(pso,shade,pop,[](std::span<const
float> x) { return rosenbrock(x); },lb,ub,alg);

std::cout << "best value:"<<
std::min_element(pop.begin(), pop.end(),[](auto &l,
auto &r) { return l.fitness() < r.fitness(); })-
>fitness()<< '\n';
```

**Listing 4: PSODE implementation in SEvoBench (pairwise/3 selection variant)**

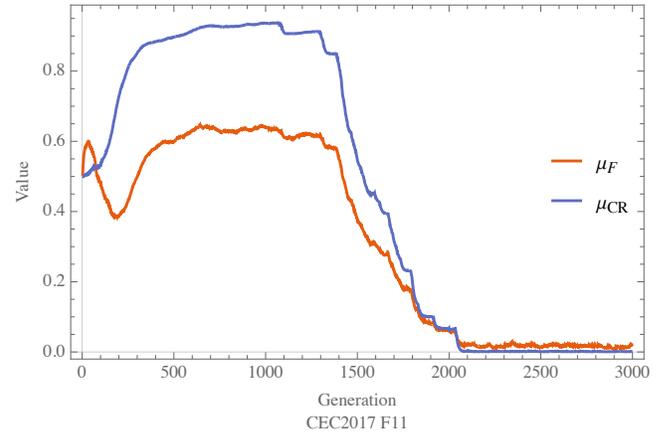

**Figure 9: Variation graphs of $\mu_F$ and $\mu_{CR}$**